\documentclass{article}

\usepackage{arxiv}

\usepackage{amsfonts}       
\usepackage{nicefrac}       
\usepackage{microtype}      
\usepackage{lipsum}		
\usepackage{natbib}
\usepackage{doi}

\usepackage{graphicx}
\usepackage[misc,geometry]{ifsym} 
\usepackage{fontawesome}
\usepackage{academicons}
\usepackage{hyperref} 
\usepackage{amsmath}
\usepackage[bottom]{footmisc}
\usepackage{supertabular}
\usepackage{afterpage}
\usepackage{url}
\usepackage{pifont}
\usepackage{multicol}
\usepackage{multirow}
\usepackage{tikz}
\usetikzlibrary[topaths]
\usetikzlibrary{arrows}
\usetikzlibrary{fit,backgrounds}
\usepackage{booktabs}
\usepackage{xcolor,colortbl}
\usepackage{color,soul}
\usepackage{blindtext}
\definecolor{lavender}{rgb}{0.9, 0.9, 0.98}
\usepackage{amssymb}
\usepackage{lscape}
\usepackage{arydshln}
\usepackage{makecell}
\usepackage{array}
\usepackage{stfloats}

\newcommand\deit{$\text{DeiT}_{ \rm BASE}$ }
\newcommand\swin{$\text{Swin}_{ \rm BASE}$ }
\newcommand\vit{$\text{ViT}_{ \rm BASE}$ }
\newcommand\bert{$\text{BERT}_{ \rm BASE}$ }
\newcommand\distilbert{$\text{DistilBERT}_{ \rm BASE}$ }
\newcommand\gpt{$\text{GPT-2}_{ \rm SMALL}$ }

\usepackage{soul}

\setcitestyle{square}

\definecolor{orcidlogo}{rgb}{0.37,0.48,0.13}
\definecolor{unilogo}{rgb}{0.16, 0.26, 0.58}
\definecolor{maillogo}{rgb}{0.58, 0.16, 0.26}
\definecolor{darkblue}{rgb}{0.0,0.0,0.0}
\hypersetup{colorlinks,breaklinks,
            linkcolor=darkblue,urlcolor=darkblue,
            anchorcolor=darkblue,citecolor=darkblue}

\title{Brazilian Portuguese Image Captioning with Transformers: A Study on Cross-Native-Translated Dataset}

\author{
    \href{https://orcid.org/0009-0000-6816-7913}{\includegraphics[scale=0.06]{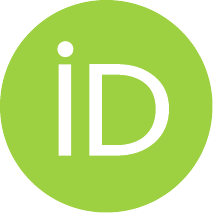}
    \hspace{1mm} Gabriel Bromonschenkel} \\
    Instituto Federal do Espírito Santo (IFES)\\
	Serra, Brazil \\
	\texttt{gabriel.mota.b.lima@gmail.com} \\
	\And
    \href{https://orcid.org/0009-0000-6816-7913}{\includegraphics[scale=0.06]{orcid.pdf}
    \hspace{1mm} Alessandro L.~Koerich} \\
    \'Ecole de Technologie Sup\'erieure (\'ETS) \\
	Montreal, Canada \\
	\texttt{alessandro.koerich@etsmtl.ca} \\
	\And
    \href{https://orcid.org/0000-0003-1554-6834}{\includegraphics[scale=0.06]{orcid.pdf}
    \hspace{1mm} Thiago M.~Paixão} \\
    Instituto Federal do Espírito Santo (IFES)\\
	Serra, Brazil \\
	\texttt{thiago.paixao@ifes.edu.br} \\
	\And
    \href{https://orcid.org/0000-0003-0643-7206}{\includegraphics[scale=0.06]{orcid.pdf}
    \hspace{1mm} Hilário Tomaz Alves de Oliveira} \\
    Instituto Federal do Espírito Santo (IFES)\\
	Serra, Brazil \\
	\texttt{hilario.oliveira@ifes.edu.br} \\
}

\renewcommand{\headeright}{}
\renewcommand{\undertitle}{}
\renewcommand{\shorttitle}{}

\hypersetup{
pdftitle={Brazilian Portuguese Image Captioning with Transformers: A Study on Cross-Native-Translated Dataset},
pdfsubject={cs.CV, cs.CL, cs.LG},
pdfauthor={Gabriel Bromonschenkel, Alessandro L.~Koerich, Thiago M.~Paixão, Hilário Tomaz Alves de Oliveira},
pdfkeywords={Image Captioning, Transformers, Brazilian Portuguese, Vision Encoder-Decoder, Multi-Modal Evaluation, Attention Maps, CLIP-Score, Vision-Language Models},
}

\begin{document}

\maketitle

\begin{abstract}
\textbf{Abstract.~}
\noindent Image captioning (IC) refers to the automatic generation of natural language descriptions for images, with applications ranging from social media content generation to assisting individuals with visual impairments. 
While most research has been focused on English-based models, low-resource languages such as Brazilian Portuguese face significant challenges due to the lack of specialized datasets and models. 
Several studies create datasets by automatically translating existing ones to mitigate resource scarcity.
This work addresses this gap by proposing a cross-native-translated evaluation of Transformer-based vision and language models for Brazilian Portuguese IC. 
We use a version of Flickr30K comprised of captions manually created by native Brazilian Portuguese speakers and compare it to a version with captions automatically translated from English to Portuguese.
The experiments include a cross-context approach, where models trained on one dataset are tested on the other to assess the translation impact. 
Additionally, we incorporate attention maps for model inference interpretation and use the CLIP-Score metric to evaluate the image-description alignment.
Our findings show that Swin-DistilBERTimbau consistently outperforms other models, demonstrating strong generalization across datasets. 
ViTucano, a Brazilian Portuguese pre-trained VLM, surpasses larger multilingual models (GPT-4o, LLaMa 3.2 Vision) in traditional text-based evaluation metrics, while GPT-4 models achieve the highest CLIP-Score, highlighting improved image-text alignment. 
Attention analysis reveals systematic biases, including gender misclassification, object enumeration errors, and spatial inconsistencies.
The datasets and the models generated and analyzed during the current study are available in: \url{https://github.com/laicsiifes/transformer-caption-ptbr}.
\end{abstract}


\section{Introduction} \label{sec:intro}

Image Captioning (IC) bridges the gap between vision and language in computational applications, enabling machines to generate coherent and meaningful descriptions of visual content. 
This interdisciplinary task holds potential for applications such as enhancing visual accessibility, aiding medical prescriptions and diagnoses, facilitating image-text indexing for information retrieval, and advancing human-computer interaction \citep{ghandi2023deep, stefanini2022show, sharma2023comprehensive}.

Advances in IC models have been driven by deep learning approaches, notably Convolutional Neural Networks (CNNs) and Transformers \citep{stefanini2022show, sharma2023comprehensive}. 
Despite significant progress in well-resourced (datasets and models) languages like English, those with limited resources, such as Brazilian Portuguese, remain underexplored.
Captions generated in English and then translated into other languages often lack the accuracy and contextual relevance achieved by models trained directly on native datasets in the target language \citep{santos2022mdpi, gondim2022towards}.

The development of robust IC models for low-resource languages depends significantly on the availability and quality of both translated and native datasets. 
Translated datasets, such as the adapted Brazilian Portuguese version of Flickr30K, enable researchers to leverage large-scale datasets from high-resource languages for cross-lingual experimentation \citep{gondim2022towards, bromonschenkel2024comparative, dos2023capivara}. 
However, translations often fail to capture cultural and linguistic nuances, potentially leading to suboptimal model performance. 
In contrast, native datasets, like \#PraCegoVer-63K, authentically represent linguistic and contextual nuances, ensuring models better align with specificities of the target language \citep{santos2022mdpi}.

A preliminary study conducted by our group \citep{bromonschenkel2024comparative} investigated Transformer-based Vision Encoder-Decoder (VED) models for IC considering both translated and native datasets.
As far as we know, it was the first comprehensive evaluation of various VED configurations for Brazilian Portuguese. 
However, the investigation was limited to a single-database evaluation, assessing performance on either translated or native datasets.
Additionally, the quantitative analysis did not incorporate cross-modal image-text metrics (e.g., CLIP-Score), and the qualitative assessment of generated captions lacked visual attention maps, which could improve model interpretability.

This study introduces a comprehensive analysis of the effects of dataset translation quality on image captioning performance by comparing an automatically translated version of Flickr30K with a manually annotated counterpart.
The research provides valuable insights into the interplay between linguistic precision and machine learning outcomes by examining how translation quality influences model performance and caption accuracy.
Additionally, we conduct extensive quantitative and qualitative evaluations to deepen our understanding of model behavior.
A key aspect of this work is the application of the CLIP-Score metric, which facilitates interpretable assessments of alignment between generated captions and reference images, surpassing traditional reference-based metrics in accessibility.
To further enhance interpretability, attention visualization is employed to reveal the image regions the model prioritizes during caption generation. This study builds upon our earlier investigation \citep{bromonschenkel2024comparative}, expanding its scope and depth to offer a more fine perspective on these critical aspects.

This research aims to address the depicted issues in Brazilian Portuguese image captioning by presenting:

\begin{itemize}
    
    \item A comprehensive evaluation of Transformer-based vision encoder-decoder models trained and evaluated on native and translated versions of the Flickr30K dataset. \\

    \item A quantitative analysis of smaller, fine-tuned VED models compared to both open-source and closed-source large pre-trained Visual Language Models (VLMs). \\
    
    \item A cross-source evaluation setup where models trained on machine-translated datasets are evaluated on native human-generated datasets and vice versa. Such analysis enables quantifying the impact of translation on model performance. \\
    
    \item The adoption of CLIP-Score within the CAPIVARA framework for cross-modal evaluation in Brazilian Portuguese, comparing its efficacy against traditional metrics such as BLEU, ROUGE, METEOR, and CIDEr. \\
   
    \item The integration of attention maps to visually interpret model decision-making in the Brazilian Portuguese domain, providing insights into the relationship between generated captions and corresponding image regions.

\end{itemize}

The remaining content is organized into five sections. 
Section~\ref{sec:related-work} presents the related work containing the historical studies focused on multimodal datasets and models for Brazilian Portuguese IC. 
Section~\ref{sec:transformers} shows the main architectural approach of this work, the vision encoder-decoder models. 
Section~\ref{sec:methods} explains the datasets, performance metrics, and experimental setup. 
Section~\ref{sec:results} presents the results obtained from experiments and explains them more deeply with qualitative analyses. 
Finally, Section~\ref{sec:conclusion} synthesizes the results and findings of our work and suggests new avenues for further research.

\section{Related Work} \label{sec:related-work}
 
Early deep learning approaches relied predominantly on CNNs for visual feature extraction and Recurrent Neural Networks (RNNs), such as Long Short-Term Memory (LSTM) networks, for sequence generation \citep{stefanini2022show, sharma2023comprehensive}. 
For example, \citet{vinyals2015show} proposed a foundational framework combining visual and language components, which paved the way for integrating attention mechanisms and Transformer-based architectures.
The introduction of Transformer-based models marked a paradigm shift in IC. 
Self-attention mechanisms enabled these models to effectively capture long-range dependencies between words and image features. 
Models such as Vision Transformer (ViT) \citep{dosovitskiy2020image} and Swin Transformer \citep{liu2021swin} achieved state-of-the-art performance on a variety of vision and multi-modal tasks. 
These models use image patches as inputs, analogous to tokens in Natural Language Processing (NLP) tasks; thereby, they can process the smaller parts of global information to extract the relations between visual elements.

In Brazilian Portuguese, \citet{gondim2022towards} evaluated a CNN-RNN with attention where the CNN block is composed of an EfficientNet of the B7 family. 
Their work has two crucial stages: dataset translation and two-step performance rating, where an automatic assessment and human evaluation are carried out. 
In the first stage, the Flickr8K dataset is translated to Portuguese using LibreTranslate. 
For the evaluation stage, performance is assessed with BLEU and METEOR, followed by an evaluation with 34 human annotators using 100 examples of captions generated by the best model.
Their study demonstrated that while translations provide a practical way to adapt high-resource datasets, they often fail to capture linguistic and contextual nuances, which limits the quality of generated captions. 
Due to the Flickr8K dataset size, the work of \citet{gondim2022towards} is constrained to a smaller quantity of information and context compared to larger datasets, such as the Flickr record with 30,000 examples.

\citet{de2024image} explored the Grid-and-Region-based Image Captioning Transformer model on a Portuguese-translated version of the MSCOCO dataset. 
This fully Transformer-based architecture demonstrated strong performance in adapting advanced IC methodologies to Brazilian Portuguese. 
However, like other works using translated datasets, it faced limitations in representing the cultural and linguistic specificity of the target language.

To address the scarcity of resources in Brazilian Portuguese, \citet{santos2022mdpi} introduced \#PraCegoVer, the first large-scale dataset for image captioning in such a language.
The dataset, sourced from Instagram posts tagged with the hashtag \#PraCegoVer, features user-provided captions aimed at promoting accessibility for visually impaired individuals. 
This dataset highlights the challenges of captioning in Brazilian Portuguese, given its cultural and linguistic diversity. 
The authors trained an Attention-on-Attention Network (AoANet) to learn from the \#PraCegoVer dataset, putting AoANet forward as the first model for image captioning trained in a native dataset for Brazilian Portuguese.

Recent studies have shown the effectiveness of the Vision Encoder-Decoder architecture in more specialized domains, including medical image captioning and multilingual captioning. 
For instance, \citet{jnaini2024synergy} explored the synergy between GPT-3 summarization and VED models for generating accurate descriptions of chest X-ray images. 
Their results indicated that the VED architecture effectively captures visual features and aligns them with textual descriptions, leading to improved captioning performance.
Similarly, \citet{abdelaal2024image} investigated the use of VED models for generic image captioning tasks, demonstrating that Transformer-based vision-language models outperform traditional CNN-RNN architectures by leveraging self-attention mechanisms for better feature extraction and text generation. 
Furthermore, \citet{ishan2023bengali} applied a VED model to Bengali image captioning, confirming its adaptability to low-resource languages. Their work highlighted that pre-trained vision encoders and language decoders enhance caption fluency and contextual accuracy, even in languages with limited labeled datasets.

Our work differs from previous research for Brazilian Portuguese IC \citep{gondim2022towards, de2024image, santos2022mdpi} in several aspects, as shown in Table~\ref{tab:comparison}. 
Unlike prior studies, which used either translated or native datasets, we employ both in Brazilian Portuguese under  Flickr30K. 
Furthermore, we evaluate multiple VED models, rather than focusing on a single architecture.
Finally, we compare the performance of large, pre-trained VLMs without fine-tuning against our fine-tuned VED models to better understand their respective strengths and trade-offs.

Additionally, we incorporate CLIP-Score within the CAPIVARA framework for reference-free evaluation in the quantitative results, alternatively to previous work, which relied only on reference-based metrics or carried out the review with human annotators in a restrictive volume of data. 
For qualitative results, we incorporate attention maps to improve the understanding of which focal points the model is inclined to use. 
By integrating these elements, our study offers a more extensive and systematic evaluation of image captioning in Brazilian Portuguese compared to previous approaches.

\begin{table*}
\renewcommand{\arraystretch}{1.3}
\centering
\caption{Comparison of Brazilian Portuguese Image Captioning Works.}
\resizebox{\linewidth}{!}{
\begin{tabular}{lcccccc}
\toprule
\textbf{Work} & 
\makecell{\textbf{Native} \\ \textbf{Dataset}} & 
\makecell{\textbf{Translated} \\ \textbf{Dataset}} & 
\makecell{\textbf{Transformer-} \\ \textbf{based} \\ \textbf{Models}} & 
\makecell{\textbf{Analyze} \\ \textbf{Translation} \\ \textbf{Impact}} & 
\makecell{\textbf{Cross-Modal} \\ \textbf{Evaluation}} & 
\makecell{\textbf{Attention} \\ \textbf{Maps} \\ \textbf{Investigation}} \\
\midrule
\citet{santos2022mdpi} & \ding{51} &  &  &  &  & \\
\citet{gondim2022towards} &  & \ding{51} &  &  &  & \\
\citet{de2024image} &  & \ding{51} & \ding{51} &  &  & \\ \hdashline
Our Work & \ding{51} & \ding{51} & \ding{51} & \ding{51} & \ding{51} & \ding{51} \\
\bottomrule
\end{tabular}
}
\label{tab:comparison}
\end{table*}

\section{Transformer-based Image Captioning} \label{sec:transformers}

Typically, IC models benefit from an encoder-decoder architecture, where the encoder processes the image to an embedding space in the format of the neural network's hidden states. 
The decoder converts these hidden states into natural language descriptions. 
The vision encoder-decoder models of this work harness checkpoints of pre-trained Transformers, like BERT.

\subsection{Visual Encoders}

We employ a Transformer-based Vision Encoder-Decoder architecture, combining visual encoders for processing image content and language decoders for generating the captions.

The three visual encoders in our Vision Encoder-Decoder configurations are state-of-the-art Transformer-based architectures for vision, which are pre-trained on ImageNet-1K at 224px resolution.
The first is Vision Transformer (ViT), a pioneering transformer-based architecture for computer vision tasks. 
Unlike convolutional networks, ViT processes images by dividing them into fixed-size patches (e.g., 16x16 pixels), which are flattened and treated as input tokens. 
Each patch is linearly embedded and augmented with positional encodings to retain spatial information. 
These tokens are then passed through multiple self-attention layers to model global dependencies between image regions.
Our work explores the ViT base version with 12 layers and 87 million parameters \citep{dosovitskiy2020image}.

The second vision encoder assessed is the Shifted Window (Swin) Transformer, a model with four stages of patch processing. 
In the first stage, 4px resolution patches are linearly embedded, followed by progressive merging in subsequent stages until the resolution reaches 32px.
In this way, Swin Transformer builds on the strengths of ViT by introducing a hierarchical representation with shifted windows. 
Images are divided into non-overlapping windows, and self-attention is applied within each window, significantly reducing computational cost compared to global attention approaches, such as ViT. 
The shifted window mechanism allows for cross-window interaction and captures both local and global image features.
This investigation uses the Swin Transformer base version with 24 layers distributed through 4 blocks separated by patch embedding layers, containing 88 million parameters \citep{liu2021swin}.

The third encoder investigated is a Data-Efficient Image Transformer (DeiT), which comprises a variant of ViT designed to achieve competitive performance with fewer data and computational resources. 
It incorporates a teacher-student knowledge distillation approach, in which a convolutional neural network acts as a teacher to improve the training efficiency of DeiT. 
Additionally, it uses a classification token to aggregate global information for image understanding.
This study adopts the DeiT base version with 12 layers and 86 million parameters \citep{touvron2021training}.

\subsection{Language Models}

The three language decoders assessed in our VED configurations are models designed for text generation and language processing in Portuguese.
The first is BERTimbau, a Bidirectional Encoder Representations from Transformers (BERT) based encoder model pre-trained on the Brazilian Web as Corpus (BrWaC) dataset. 
It uses a masked language modeling (MLM) objective during pre-training, where a chunk of the input tokens are masked, and the model learns to predict them based on context. 
Our research examines the base version with 12 layers and 110 million parameters \citep{souza2020bertimbau}.

The second is DistilBERTimbau, which is a distilled version of BERTimbau-based designed to reduce model size and inference latency while retaining most of its performance. 
The distillation process involves training the smaller model to mimic the outputs of a larger BERTimbau model. 
This results in a lightweight model with a fraction of the computational requirements of BERTimbau. 
For this work, we adopt the base version with 6 layers and 66 million parameters \citep{silva2022towards, adalberto_ferreira_barbosa_junior_2024}.

GPorTuguese-2 is the third model evaluated. 
It is a Generative Pre-trained Transformer (GPT)-based model derived from a fine-tuned small version of GPT-2 on Portuguese Wikipedia. 
The model employs a decoder-only transformer architecture with causal attention, enabling autoregressive text generation. 
GPorTuguese-2 consists of 12 layers and 137 million parameters \citep{pierre2020gpt2smallportuguese}.

\begin{figure*}[ht]
    \centering    
    \includegraphics[width=0.9\linewidth]{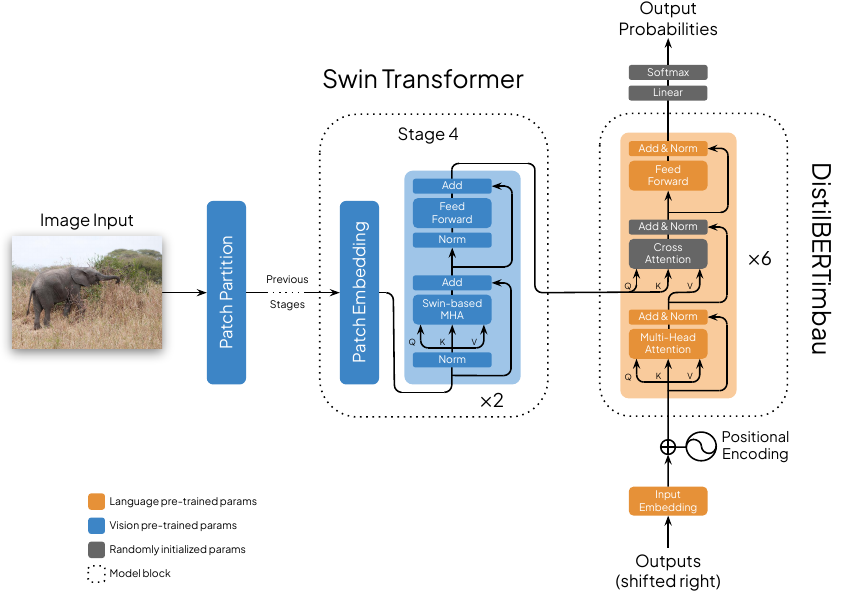}
    \caption{Illustration of the architecture on the merge of Swin Transformer and DistilBERTimbau. Swin-DistilBERTimbau is one of the nine VED combinations used in this work. MHA stands for Multi-Head Attention.}
    \label{fig:illustration-swin-distilbertimbau}
\end{figure*}

To visually introduce the vision encoder-decoder design, Figure~\ref{fig:illustration-swin-distilbertimbau} shows one out of the nine architectural combinations of our work, using the model resulting from the merger between Swin Transformer and DistilBERTimbau, the coupling with the best performance in Table~\ref{tab:same-data}. 
The visual encoders generate feature embeddings that capture the spatial and semantic information of the image.
These embeddings are then projected as 768-dimensional tensors into the language decoders, which align them with textual representations to produce captions.
The alignment process is boosted by cross-attention, positional encodings, and fine-tuning.
Through cross-attention, the decoders focus on specific regions of the image embeddings, ensuring that the generated captions are contextually grounded in the visual input. 
Both encoders and decoders use positional encodings to preserve spatial and sequential information, which is crucial for modeling the relationships between objects in images and words in captions. 
The models are fully fine-tuned on either native or translated datasets, ensuring they adapt to the linguistic and contextual nuances of Brazilian Portuguese.

\section{Experimental Methodology} \label{sec:methods}

This section presents the datasets, performance metrics, and experiments. First, the datasets are introduced, considering both translated and native contexts while highlighting data curation nuances and dataset characteristics. Next, we briefly explain the metrics used for performance evaluation. In the following, we describe the experiments for both quantitative and qualitative evaluation, with the section concluding with implementation details of the experiments.  

\subsection{Datasets} \label{sec:datasets}

This study uses two datasets derived from the Flickr30K dataset, each providing a different source of image-caption pairs in Brazilian Portuguese. 
One dataset consists of human-generated captions, while the other contains machine-translated captions from English.

The first dataset, Flickr30K in Native Portuguese (Flickr-Native), is based on the FM30K corpus \citep{viridiano2024framed}, which in turn is an extension of the Flickr30K dataset, embodying more examples. 
It contains 31,014 images, each associated with five captions written by native Portuguese speakers. 
The dataset maintains the original image-caption pairs from Flickr30K but includes manually created Portuguese descriptions. 
These captions follow the linguistic structures of the language without translation artifacts. 
The FM30K is designed to be a multi-modal and multi-lingual dataset.
It expands Multi30K (a German extension of Flickr30K) with Brazilian Portuguese descriptions and frame resources.

The second dataset, Flickr30K Translated to Portuguese (Flickr-Translated), consists of the same 31,014 images, but the original captions were translated from English using Google Translate\footnote{\url{https://translate.google.com/}}. 
This dataset allows an evaluation of the effects of translation on image captioning models. 
Automatic translations can introduce errors such as unnatural phrasing and altered word order, which may affect model performance.

Table~\ref{tab:datasets} summarizes statistics for both native and translated datasets. 
The statistical information is generated using the SpaCy toolkit\footnote{\url{https://spacy.io/}}, configured to ignore punctuation.
On average, captions in Flickr-Native are longer than those in Flickr-Translated, suggesting that manually created descriptions tend to be more detailed.
The toolkit identifies 20,719 distinct words in Flickr-Translated, of which 9,412 are absent in the native source. 
In contrast, Flickr-Native contains 18,990 distinct words, with 7,683 words not found in the translated source.

\begin{table}[hb]
    \centering
    \setlength{\tabcolsep}{4pt}
    \renewcommand{\arraystretch}{1.3}
    \caption{Descriptive statistics of the datasets' splits. Val. is an abbreviation for Validation.}
    \label{tab:datasets}
    \begin{tabular}{lccc}
        \toprule
        &  & \multicolumn{2}{c}{Avg. Caption Length (\#Words)}  \\
        \cmidrule(rl){3-4}
        \textbf{Split} & \textbf{\#Examples}& \textbf{Flickr-Translated} & \textbf{Flickr-Native} \\
        \midrule
        Train & 29,000 & 12.1 $\pm$ 5.1 & 13.4 $\pm$ 5.4 \\
        Val. & 1,014 & 12.3 $\pm$ 5.3 & 13.5 $\pm$ 5.5 \\
        Test & 1,000 & 12.2 $\pm$ 5.4 & 13.4 $\pm$ 5.4 \\
        \hdashline
        Total & 31,014 & 12.1 $\pm$ 5.2 & 13.4 $\pm$ 5.4 \\
        \bottomrule
    \end{tabular}
\end{table}

\subsection{Performance Metrics}

To assess the generated captions, we incorporated CLIP-Score in addition to the reference-based (uni-modal) metrics used in our preliminary study. 
CLIP-Score is a model-based multi-modal metric that quantifies the similarity between image and text content within a shared embedding space. 
Therefore, the metrics computed in our experiments are:

\begin{itemize}
    
    \item \textbf{CIDEr-D}: A consensus-based metric that weights n-grams based on their Term Frequency-Inverse Document Frequency (TF-IDF) scores across reference captions.
    It is specifically developed for description evaluation in image captioning task \citep{vedantam2015cider}. \\
   
    \item \textbf{BLEU-4}: A precision-based metric that evaluates n-gram overlaps up to 4-grams between generated and reference captions. 
    It was initially designed to measure the text quality in automatic translation task \citep{papineni2002bleu}. \\
    
    \item \textbf{ROUGE-L}: A metric based on the longest common subsequence between generated and reference captions. 
    It is part of the Recall-Oriented Understudy for Gisting Evaluation (ROUGE) group of measures initially drawn to quantify text summarization quality \citep{lin2004rouge}. \\
    
    \item \textbf{METEOR}: The Metric for Evaluation of Translation with Explicit Ordering (METEOR) is a recall-oriented metric that considers synonyms and word stemming for improved linguistic evaluation. 
    It is typically used for text translation assessment \citep{banerjee2005meteor}. \\
   
    \item \textbf{BERTScore}: A semantic similarity metric using contextual embeddings generated by a BERT-based model. 
    We use the F1-score derived from BERTScore, using the BERTimbau as the embedder for Portuguese captions \citep{zhang2019bertscore}. \\
    
    \item \textbf{CLIP-Score}: A cross-modal metric that measures the similarity between image embeddings and text embeddings using a model based on Contrastive Language-Image Pre-training (CLIP) \citep{hessel2021clipscore}. 
    We leverage the CAPIVARA framework \citep{dos2023capivara} to adapt CLIP-Score’s performance for Brazilian Portuguese.

\end{itemize}

Both the reference-based (uni-modal) metrics and CLIP-Score have the potential to quantify the accuracy of generated captions. 
While CLIP-Score captures terms and points of view not depicted in the reference captions, the reference-based metrics maintain the fidelity of the analysis anchored in the reference captions. 
On the other hand, as reported by CLIP-Score authors, the multi-modal evaluation is attached to the model pre-trained bias. 
The bias problem can insert erroneous judgment into the candidate caption assessment. 
For conciseness, BLEU-4, ROUGE-L, METEOR, BERTScore, and CLIP-Score will be shortened to B@4, RL, M, BS, and CS. 
The BLEU, ROUGE, METEOR, BERTScore, and CLIP-Score metrics range from 0 to 1, while CIDEr-D can exceed this range.

\subsection{Experiments}

\begin{figure}[hb]
    \centering
    \includegraphics[width=.55\columnwidth]{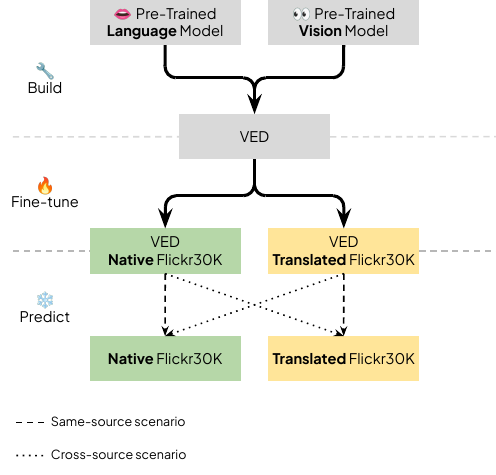}
    \caption{Simplified illustration of the experiment pipeline.}
    \label{fig:experiment-roadmap}
\end{figure}

\paragraph{VED Models Assessment.} The first experiment aims to comprehensively evaluate the VED models in both same-source and cross-source setups, as depicted in Figure~\ref{fig:experiment-roadmap}. In the same-source setup, the goal is to establish the conventional performance of the models by training and testing within the same setting (Native-to-Native and Translated-to-Translated). Cross-source evaluation assesses context shifting by considering Native-to-Translated and Translated-to-Native scenarios, allowing for an analysis of how well models generalize across human-generated and machine-translated captions. 
This assessment is particularly useful for quantifying the source-drift impact of training models on the translated dataset and applying them to a native dataset (Translated-to-Native), as it reduces the human effort required to produce large-scale training data in the native language. 

\paragraph{VLMs Assessment.} For a more comprehensive analysis, state-of-the-art Vision-Language Models (VLMs) were evaluated in a zero-shot setting to establish baseline performance. The fine-tuned VED models were compared with the following pre-trained VLMs (not fine-tuned on the datasets used in this study): the open-source ViTucano 1B and 2B \citep{Correa20204VITucano}, PaliGemma \citep{Beyer2024Paligemma}, Phi-3 Vision (Phi-3 V) \citep{Abdin2024Phi3}, and LLaMa3.2 Vision 11B (LLaMa3.2 V) \citep{Dubey2024Llama}, alongside the proprietary models GPT-4o and GPT-4o-mini \citep{hurst2024gpt}.
The ViTucano models are natively pre-trained for Brazilian Portuguese, while the others are multilingual.
Caption generation followed the prompt: ``Escreva uma descrição em português do Brasil para a imagem com no máximo 25 palavras.'' (``Write a description in Brazilian Portuguese for the image with a maximum of 25 words.''), except for PaliGemma, which used the author-reported prompt ``caption pt''. 
Performance metrics were computed by comparing generated captions with reference captions from both Flickr30K Translated and Flickr30K Native, using the same 1,000-image test split.
PaliGemma, Phi-3V, ViTucano 1B and 2B are used through  Transformers library\footnote{\url{https://huggingface.co/transformers}}, while LLaMa3.2 V is utilized via Unsloth library\footnote{\url{https://unsloth.ai/}}. PaliGemma, Phi-3 V, and LLaMa3.2 V are compressed with 4-bit quantization to fit within the available resources.
The GPT-4o models are accessed via the OpenAI API\footnote{\url{https://platform.openai.com/docs/overview}}.

\paragraph{Qualitative Assessment.} Swin-DistilBERTimbau, the model with the highest performance across most metrics in the same-source setup, was selected for a more in-depth qualitative assessment. For this analysis, we selected examples from both Native-to-Native and Translated-to-Translated scenarios, focusing on cases with (comparatively) high performance on CLIP-Scores in view of the remaining metrics and vice versa. Additionally, we analyze the quality of the captions by examining the attention maps for the Swin-DistilBERTimbau. 

\subsection{Implementation Details}

\paragraph{Training the Models.} All VED models are trained for 20 epochs on an NVIDIA RTX 4090 GPU, 24GB VRAM. To improve caption generation, we use the Adam optimizer with a learning rate of 5e-5, a batch size of 16, and a beam search decoding strategy with a beam size of 5. Pre-trained checkpoints for both vision encoders (ViT, Swin, DeiT) and language decoders (BERTimbau, DistilBERTimbau, GPorTuguese-2) are leveraged for transfer learning, ensuring stable convergence and improved performance.

\paragraph{Attention Maps.} The implementation of the attention maps is inspired on \citet{liu2021cptr}\footnote{A similar implementation can be found in the TensorFlow tutorial at \url{https://www.tensorflow.org/text/tutorials/image_captioning\#attention_plots}}. The cross-attention layer is the architectural element that associates vision and language modalities, while the other multi-head layers are self-attentive, implementing uni-modal attention. Similarly, \citet{liu2021cptr} uses the token-patches attention abstracted in the cross-attention layer.  In our study, we use the cross-attention layer of the last decoder block, which is the output block directly generating the captions. Different VED architectures have peculiarities in the internal attention states that should be regarded when computing attention maps. For instance, the Swin Transformer (the encoder in Swin-DistilBERTimbau) employs a strategy of increasing patch size through image block processing, starting with a patch size of 4 and ending with a patch size of 32 (i.e., from a grid of 56$\times$56 to 7$\times$7 patches). In the cross-attention layer, this architectural structure results in tensors of attention's alphas from 12 attention heads of DistilBERTimbau and 7$\times$7 patches of Swin Transformer.

\section{Results and Discussion} \label{sec:results}

This section presents and discusses the results of the conducted experiments.

\subsection{VED Models Assessment} \label{sec:veds-results}

\subsubsection{Same-source Setup}

Table~\ref{tab:same-data} shows the VED model's performance for the same-source setup on both translated and native versions of Flickr30K. In both cases, we observe that Swin-based encoders systematically outperform DeiT and ViT encoders in most metrics. Out of Swin-based models, the only models that find a place among the three best models are the GPT-based models for the native context.

For Flickr30K Translated, Swin-DistilBERTimbau achieved the highest performance in uni-modal metrics, while Swin-BERTimbau yielded a slightly higher CLIP-Score (53.81). 
On Flickr30K Native, Swin-GPorTuguese-2 dominates CIDEr, ROUGE, BLEU, and METEOR, outperforming the other encoder-decoder combinations.
On the other hand, Swin-DistilBERTimbau achieved the highest values in model-based metrics, i.e., BERTScore and CLIP-Score.
Model-based metrics (i.e., BERTScore, CLIP-Score) provide a better image and language context assessment of the generated captions compared to relying solely on lexical matching.
Thus, these metric results indicate that for the native context, Swin-GPorTuguese-2 achieved a better lexical match with the reference captions, while Swin-DistilBERTimbau had a better contextual match.

Moreover, we noticed that the models generally performed better on machine-translated data (compared to native text) for most lexical metrics.
The only performance peculiarity is revealed by GPT-based models in Native-to-Native, which achieved ROUGE-L and BLEU-4 higher than its equivalents in the Translated-to-Translated scenario.
This gap underscores that the Flickr-Native is more challenging than Flickr-Translated in the same-dataset evaluation situation.
Nonetheless, the best-performing VED configurations (Swin + DistilBERTimbau/GPorTuguese-2) showed less impact across translated and native scenarios, indicating that robust visual features and carefully pre-trained Portuguese decoders are critical to achieving accurate and fluent captions.

\begin{table*}[t]
    \setlength{\tabcolsep}{4pt}
    \renewcommand{\arraystretch}{1.3}
    \caption{VED models performance (\%) for the same-source setup: Translated-to-Translated and Native-to-Native scenarios.
    The three highest values for each metric are highlighted in \textbf{bold}, with the cell corresponding to the highest value further highlighted in \colorbox{lavender}{\textbf{blue}}. Metrics: CIDEr-D (C), BLEU-4 (B@4), ROUGE-L (RL), METEOR (M), BERTScore (BS), and CLIP-Score (CS).}
    \label{tab:same-data}
    \centering
    \resizebox{\linewidth}{!}{
    \begin{tabular}{llcccccc|ccccccc}
    \toprule
    \multicolumn{2}{l}{} & \multicolumn{6}{c}{Translated-to-Translated} & \multicolumn{6}{c}{Native-to-Native} \\
\cmidrule(rl){3-8} \cmidrule(rl){9-14}
    \textbf{Encoder} & \textbf{Decoder} & \textbf{C} & \textbf{B@4} & \textbf{RL} & \textbf{M} & \textbf{BS} & \textbf{CS}  & \textbf{C} & \textbf{B@4} & \textbf{RL} & \textbf{M} & \textbf{BS} & \textbf{CS} \\
    \midrule
                        & \bert & 49.53 & 19.20 & 36.00 & 39.80 & 69.58 & 49.75 
                        & 47.84 & 17.01 & 34.01 & 39.16 & 69.71 & 50.66 \\
    \deit               & \distilbert & 50.58 & 19.24 & 35.77 & 39.93 & 69.50 & 49.67 
                        & 50.55 & 18.08 & 34.92 & 40.86 & 70.10 & 49.95 \\
                        & \gpt & 50.61 & 19.83 & 36.30 & 40.52 & 69.66 & 49.49
                        & 48.79 & 21.88 & \textbf{44.92} & 40.00 & 69.03 & 50.51 \\
                        \hdashline
                        
                        & \bert & \textbf{62.42} & \textbf{22.78} & \textbf{38.71} & \textbf{43.47} & \textbf{71.19} & \cellcolor{lavender}\textbf{53.81}
                        & \textbf{61.14} & 20.58 & 37.62 & \textbf{43.85} & \textbf{72.01} & \textbf{53.01} \\
    \swin               & \distilbert & \cellcolor{lavender}\textbf{66.73} & \cellcolor{lavender}\textbf{24.65} & \cellcolor{lavender}\textbf{39.98} & \cellcolor{lavender}\textbf{44.71} & \cellcolor{lavender}\textbf{72.30} & \textbf{53.26} 
                        & \textbf{63.77} & \textbf{22.79} & 38.06 & \textbf{44.65} & \cellcolor{lavender}\textbf{72.23} & \cellcolor{lavender}\textbf{53.26} \\
                        & \gpt & \textbf{64.71} & \textbf{23.15} & \textbf{39.39} & \textbf{44.36} & \textbf{71.70} & \textbf{53.49} 
                        & \cellcolor{lavender}\textbf{65.79} & \cellcolor{lavender}\textbf{29.17} & \cellcolor{lavender}\textbf{50.23} & \cellcolor{lavender}\textbf{45.04} & \textbf{72.06} & \textbf{53.19} \\
                        \hdashline

                        & \bert & 57.32 & 22.12 & 37.50 & 41.72 & 70.63 & 51.93
                        & 53.66 & 18.15 & 34.87 & 41.38 & 70.40 & 52.84 \\
    \vit                & \distilbert & 59.32 & 21.19 & 37.74 & 42.70 & 71.15 & 51.84
                        & 56.97 & 19.49 & 36.36 & 43.08 & 71.38 & 51.86 \\
                        & \gpt & 59.02 & 21.39 & 37.68 & 42.64 & 71.03 & 52.44
                        & 54.17 & \textbf{24.74} & \textbf{47.57} & 42.59 & 70.53 & 51.47 \\
    \bottomrule
    \end{tabular}
    }
\end{table*}

\subsubsection{Cross-source Setup}

In the cross-source setup, the models are evaluated on their ability to generalize across native and machine-translated data. Table~\ref{tab:cross-testing} shows the results for the two scenarios under the cross-source setup: (\textit{i})~Translated-to-Native (left side), where the model trained on the translated context is applied to the native context, and (\textit{ii})~Native-to-Translated (right side), the counterpart.

\begin{table*}[t]
    \setlength{\tabcolsep}{4pt}
    \renewcommand{\arraystretch}{1.3}
    \caption{VED models performance (\%) for the cross-source setup: Translated-to-Native and Native-to-Translated scenarios. The three highest values for each metric are highlighted in \textbf{bold}, with the cell corresponding to the highest value further highlighted in \colorbox{lavender}{\textbf{blue}}. Metrics: CIDEr-D (C), BLEU-4 (B@4), ROUGE-L (RL), METEOR (M), BERTScore (BS), and CLIP-Score (CS).}
    \label{tab:cross-testing}
    \centering
    \resizebox{\linewidth}{!}{
    \begin{tabular}{llcccccc|ccccccc}
    \toprule
    \multicolumn{2}{l}{} & \multicolumn{6}{c}{Translated-to-Native} & \multicolumn{6}{c}{Native-to-Translated} \\
\cmidrule(rl){3-8} \cmidrule(rl){9-14}
    \textbf{Encoder} & \textbf{Decoder} & \textbf{C} & \textbf{B@4} & \textbf{RL} & \textbf{M} & \textbf{BS} & \textbf{CS}  & \textbf{C} & \textbf{B@4} & \textbf{RL} & \textbf{M} & \textbf{BS} & \textbf{CS} \\
    \midrule
                        & \bert & 37.35 & 14.95 & 35.45 & 36.63 & 67.73 & 49.75
                        & 37.57 & 14.02 & 34.58 & 35.43 & 67.02 & 50.66 \\  
    \deit               & \distilbert & 37.52 & 14.87 & 35.28 & 36.27 & 67.55 & 49.67 
                        & 38.97 & 14.07 & 35.83 & 36.08 & 67.34 & 49.95 \\
                        & \gpt & 36.02 & 14.90 & 34.80 & 36.09 & 67.31 & 49.49 
                        & 37.58 & 11.87 & 32.58 & 35.56 & 66.64 & 50.51 \\
                        \hdashline
                        
                        & \bert & \textbf{45.37} & \textbf{17.46} & 36.90 & 37.90 & 68.57 & \cellcolor{lavender}{\textbf{53.81}} 
                        & \textbf{46.99} & \textbf{16.99} & \textbf{38.25} & \textbf{38.37} & \cellcolor{lavender}\textbf{69.19} & \textbf{53.01} \\
    \swin               & \distilbert & \cellcolor{lavender}\textbf{48.32} & \cellcolor{lavender}\textbf{18.17} & \cellcolor{lavender}\textbf{38.17} & \textbf{39.34} & \cellcolor{lavender}\textbf{69.51} & \textbf{53.26} 
                        & \cellcolor{lavender}\textbf{47.54} & \cellcolor{lavender}\textbf{17.16} & \cellcolor{lavender}\textbf{38.41} & \textbf{39.18} & \textbf{69.17} & \cellcolor{lavender}\textbf{53.26} \\
                        & \gpt & \textbf{45.41} & \textbf{17.21} & \textbf{37.75} & \cellcolor{lavender}\textbf{39.55} & \textbf{69.04} & \textbf{53.49} 
                        & \textbf{47.25} & 14.54 & 35.82 & \cellcolor{lavender}\textbf{39.20} & \textbf{68.80} & \textbf{53.19} \\
                        \hdashline

                        & \bert & 40.02 & 16.38 & 36.17 & 36.98 & 68.10 & 51.93 
                        & 42.12 & 14.50 & 35.08 & 35.71 & 67.73 & 52.84 \\
    \vit                & \distilbert & 43.45 & 16.82 & \textbf{37.46} & \textbf{38.98} & \textbf{69.08} & 51.84 
                        & 44.07 & \textbf{15.15} & \textbf{37.13} & 37.52 & 68.42 & 51.86 \\
                        %
                        & \gpt & 42.33 & 15.64 & 36.47 & 38.01 & 68.41 & 52.44 
                        & 40.43 & 12.37 & 33.45 & 36.77 & 67.53 & 51.47 \\
    \bottomrule
    \end{tabular}
    }
\end{table*}

The observed performance drop compared to the same-source setup highlights the challenges of domain shift. Nonetheless, Swin-based models achieved the highest evaluation metrics, with Swin-DistilBERTimbau assuming leadership performance across most metrics. ViT-DistilBERTimbau is a special case that achieved comparatively relevant metric estimation in ROUGE, METEOR, and BERTScore.
It is noteworthy to mention that the two contexts have an information drift due to the terms and expressions that are exclusive of each dataset, as reported in Section~\ref{sec:datasets}.

\begin{figure}[ht]
    \centering
    \includegraphics[width=.7\linewidth]{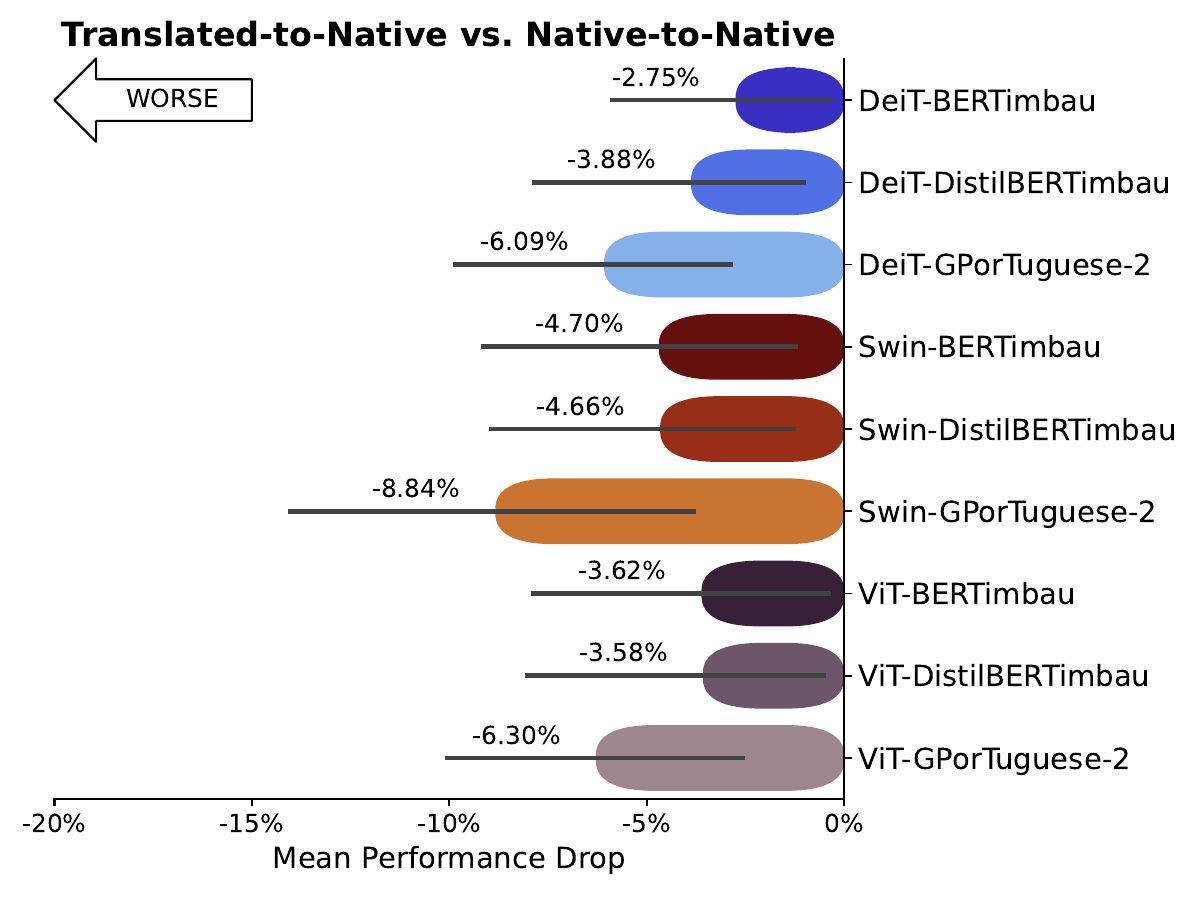}
    \caption{Illustration of the mean performance drop when switching context, using a bar chart.
    The chart depicts the mean percentage performance drop through the evaluation metrics.
    For instance, ``\textbf{Translated-to-Native vs. Native-to-Native}'' presents the difference between the Native evaluation of models trained on Translated and Native datasets. 
    The numbers over the bars are their mean percentages, and the lines in the middle of the bars are the standard deviations of the mean percentages.
    A bar color represents each model to favor the visualization.}
    \label{fig:performance-drop}
\end{figure}

\paragraph{Translated-to-Native vs. Native-to-Native}

Figure~\ref{fig:performance-drop} depicts the mean performance drop of the models tested in Flickr-Native when the training source changes from native to translated.
In this circumstance, GPorTuguese-2 models suffer more deterioration in performance than the remaining models when the training base changes from native to translated, achieving decline ranges between -8.84\% and -6.09\%, while the other models between -4.66\% and -2.75\%.

Comparing the metrics in Tables~\ref{tab:same-data} and~\ref{tab:cross-testing}, CIDEr presented the most aggressive performance drop compared to the remaining metrics. At the same time, CLIP-Score presented no performance dropping for Swin-based models and ViT-GPorTuguese-2, indicating that the quality of image-text embedding alignment is comparable to the captions generated by the models trained on the Flickr-Native. METEOR and BERTScore had a regular performance drop slightly higher than CIDEr-D. ROUGE and BLEU had similar performance drops as METEOR and BERTScore except for the GPorTuguese-2-based models, which presented a more aggressive performance drop compared to the other models.

\subsection{VLMs Assessment} \label{sec:vlms-results}

\begin{table*}[t]
    \setlength{\tabcolsep}{4pt}
    \renewcommand{\arraystretch}{1.3}
    \caption{VLM performance (\%) for zero-shot baseline VLMs on Flickr30K Translated and Flickr30K Native. The column \#Params indicates the number of model parameters in billions (B) of parameters. The three highest values for each metric are highlighted in \textbf{bold}, with the cell corresponding to the highest value further highlighted in \colorbox{lavender}{\textbf{blue}}. Metrics: CIDEr-D (C), BLEU-4 (B@4), ROUGE-L (RL), METEOR (M), BERTScore (BS), and CLIP-Score (CS). \textbf{Note}: The CS columns share the same values since this metric depends solely on the generated captions and image content, rather than the nature of the reference captions (translated or native).}
    \label{tab:baseline-vlms}
    \centering
    \resizebox{\linewidth}{!}{
    \begin{tabular}{lrcccccc|ccccccc}
    \toprule
    \multicolumn{2}{l}{} & \multicolumn{6}{c}{Pre-trained-to-Translated} & \multicolumn{6}{c}{Pre-trained-to-Native} \\
\cmidrule(rl){3-8} \cmidrule(rl){9-14}
\textbf{VLM} & \textbf{\#Params} & \textbf{C} & \textbf{B@4} & \textbf{RL} & \textbf{M} & \textbf{BS} & \textbf{CS}  & \textbf{C} & \textbf{B@4} & \textbf{RL} & \textbf{M} & \textbf{BS} & \textbf{CS} \\
    \midrule
    ViTucano 1B  & 1.53B   & \textbf{57.96} & \textbf{19.47} & \textbf{40.80} & \textbf{47.01} & \textbf{70.09} & 55.96
                        & \textbf{51.45} & \textbf{16.16} & \cellcolor{lavender}\textbf{37.96} & \cellcolor{lavender}\textbf{43.73} & \cellcolor{lavender}\textbf{69.04} & 55.96 \\
                        
    ViTucano 2B & 2.88B   & \cellcolor{lavender}\textbf{62.03} & \cellcolor{lavender}\textbf{19.84} & \cellcolor{lavender}\textbf{41.51} & \cellcolor{lavender}\textbf{47.05} & \cellcolor{lavender}\textbf{70.30} & 56.28 
                        & \cellcolor{lavender}\textbf{52.49} & \cellcolor{lavender}\textbf{16.46} & \textbf{37.75} & \textbf{43.47} & \textbf{68.87} & 56.28 \\
                        
                        \hdashline
    PaliGemma  & 2.92B     & 23.47 & 4.88 & 20.76 & 20.33 & 52.37 & 49.67
                        & 18.69 & 4.65 & 20.59 & 19.78 & 52.15 & 49.67 \\
                        
                        \hdashline

        Phi-3 V     & 4.15B  & 22.67 & 7.17 & 27.35 & 29.50 & 58.44 & 52.36 
                        & 21.92 & 7.34 & 27.94 & 30.29 & 58.81 & 52.36 \\
                        
                        \hdashline

    LLaMa3.2 V & 11.70B   & \textbf{34.48} & \textbf{10.20} & \textbf{31.56} & 35.17 & 63.18 & \textbf{56.95} 
                        & \textbf{34.94} & \textbf{9.81} & \textbf{30.81} & 34.50 & 62.84 & \textbf{56.95} \\

                        \hdashline

    GPT-4o-mini & -      & 21.68 & 8.96 & 29.31 & \textbf{41.63} & \textbf{63.83} & \textbf{61.26}
                        & 27.43 & 9.04 & 29.29 & \textbf{41.96} & \textbf{64.24} & \textbf{61.26} \\
                        
    GPT-4o      & -      & 25.68 & 7.36 & 25.04 & 38.67 & 61.49 & \cellcolor{lavender}\textbf{62.07}
                        & 34.02 & 8.93 & 27.36 & 41.18 & 63.35 & \cellcolor{lavender}\textbf{62.07} \\
    \bottomrule
    \end{tabular}
    }
\end{table*}

Table~\ref{tab:baseline-vlms} shows the results of the VLM assessment. As seen in the table, each model demonstrates unique strengths across uni-modal metrics (e.g., BLEU, CIDEr) and the multi-modal metric (CLIP-Score).
For instance, ViTucano 2B demonstrates the highest CIDEr, BLEU, ROUGE, METEOR, and BERTScore in the translated dataset and the highest CIDEr and BLEU in the native dataset. The ViTucano 1B outperforms the baseline models in ROUGE, METEOR, and BERTScore in the native context. Despite the lower performance on most text-centric metrics, GPT-4 models achieved the highest CLIP-Score, indicating strong alignment between image features and caption representations. In summary, the results show that the native pre-trained models generalized better to the translation context than in the original idiom context, and the CLIP-Score tends to increase. In contrast, the multilingual models' complexity increases, and GPT-4o was outperformed by its smaller version in all metrics except in CIDEr-D and CLIP-Score.

GPT-4o and GPT-4o-mini generated captions with an average length superior to 20 words, while the average length for PaliGemma and Phi-3 V were 8.63 and 11.39, respectively.
The other models generated captions with an average length between 14.41 and 16.23.
The GPT-4o and GPT-4o-mini showed the highest number of distinct words, 2,524 and 2,495, respectively, confirming their heterogeneous vocabulary.
The remaining models did not exceed the 2,000-word limit for distinct words.
On average, captions generated by GPT-4 models for Flickr-Translated contained more than 10 words not used in the respective reference, whereas for Flickr-Native, this number dropped to approximately 9 words. 
For both translated and native datasets, PaliGemma, Phi-3 V, and ViTucano models maintained a distinct word rate per caption of around 5, while LLaMa3.2 V averaged 6.
We observed that captions generated by the GPT-4 family are typically descriptive and concise, with a more diverse vocabulary in some examples, contradicting the reference-based metrics.

Comparing model-based metrics such as BERTScore and CLIP-Score while considering the models' vocabulary sizes highlights the limitations of evaluation metrics in capturing the coherence, conciseness, and correctness of generated captions. 
The fact that GPT-4 models use words not present in the reference captions affects uni-modal metrics. 
BERTScore, while robust to synonyms and paraphrasing due to its use of text embeddings, may fail to recognize descriptions of situations, scenes, concepts, or entities that do not exist in the reference captions. 
In contrast, CLIP-Score remains unaffected by vocabulary differences, as it relies on the alignment of extracted visual and language features.

\paragraph{VED Models vs. VLMs.} This analysis compares VED models trained and tested on native data (Native-to-Native) with pre-trained VLMs evaluated on the same native data. A comparison between Tables~\ref{tab:same-data} and \ref{tab:baseline-vlms} reveals that Swin-based models generally outperform VLMs in uni-modal metrics, underscoring the importance of fine-tuning in capturing both vocabulary and the expected textual structure of captions. However, it is worth noting that Swin-based models surpass only the Phi-3V and PaliGemma VLMs in terms of CLIP-Score. These results indicate that captions generated by VLMs -- especially the largest ones -- tend to be more aligned with image content. Nonetheless, the computational cost of such models should be considered for real-world applications. Excluding the very large GPT-4o VLMs, the most notable performance difference based on CLIP-Score is under 4 percentage points, observed when comparing the 11.70B-size VLM LLaMa3.2 V with the Swin-based models.

\subsection{Qualitative Assessment} \label{sec:clipscore}

\subsubsection{Comparing Reference and Generated Captions} 

\begin{figure}[ht]
    \centering
    \includegraphics[width=.55\columnwidth]{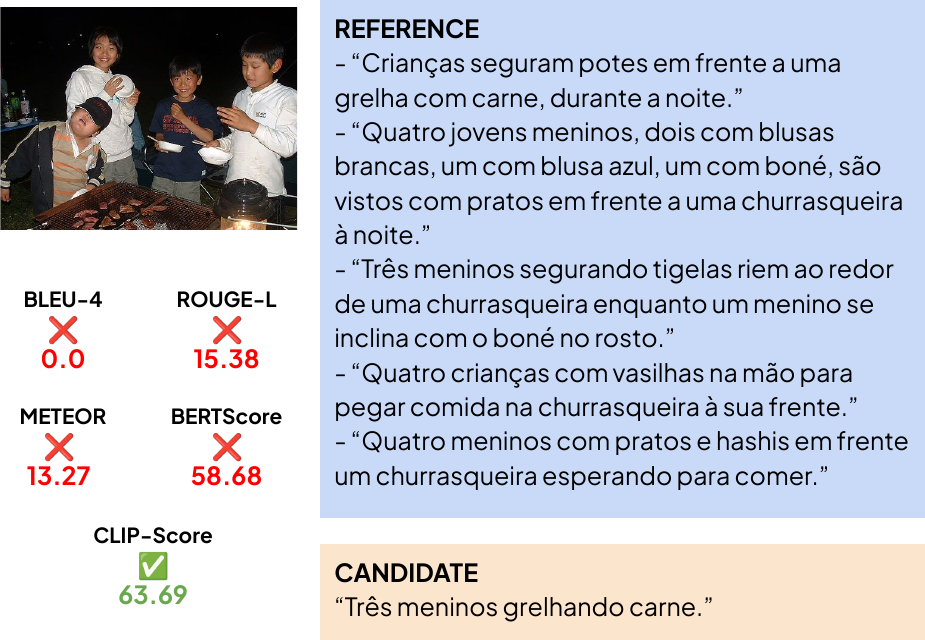}
    \caption{Case 1: Example from Flickr30K Translated with high performance on CLIP-Score compared to the uni-modal metrics.}
    \label{fig:best-clip-translated}
\end{figure}

\begin{figure}[bh]
    \centering
    \includegraphics[width=.55\columnwidth]{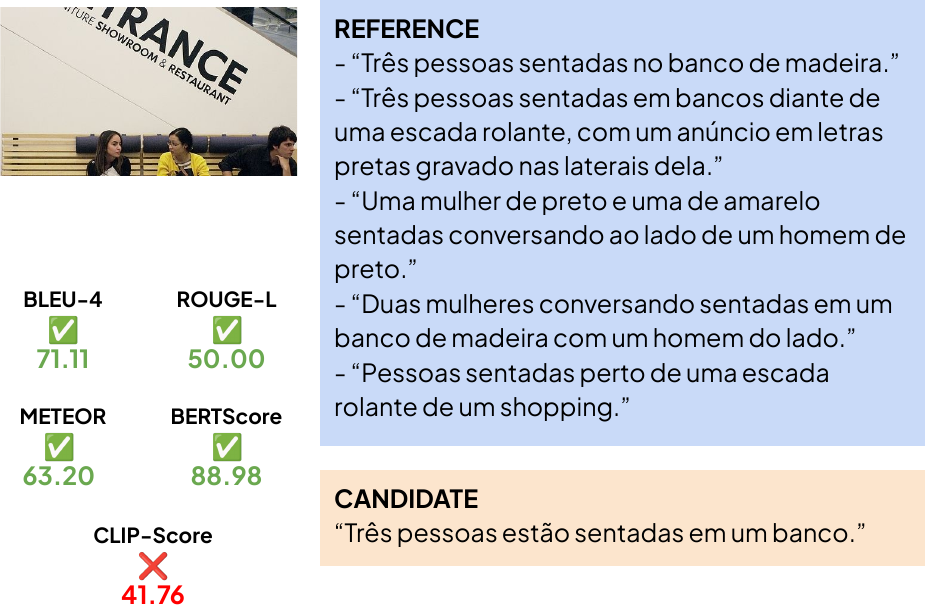}
    \caption{Case 2: Example from Flickr30K Translated with poor performance on CLIP-Score compared to the uni-modal metrics.}
    \label{fig:worst-clip-translated}
\end{figure}

\begin{figure}[ht]
    \centering
    \includegraphics[width=.55\columnwidth]{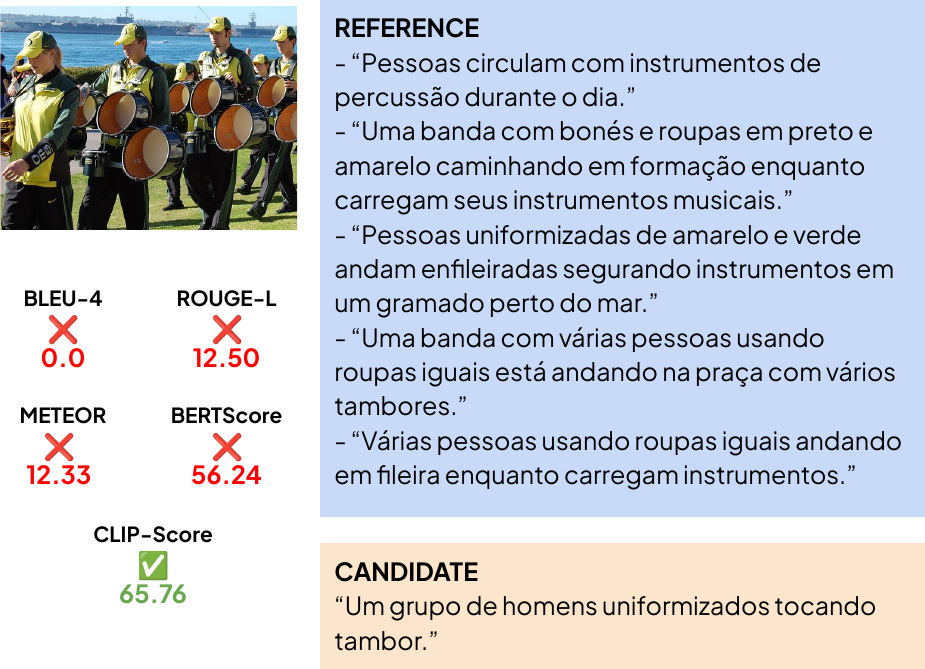}
    \caption{Case 3: Example from Flickr30k Native with a higher CLIP-Score value compared to the uni-modal metrics.}
    \label{fig:best-clip-native}
\end{figure}

\begin{figure}[ht]
    \centering
    \includegraphics[width=.55\columnwidth]{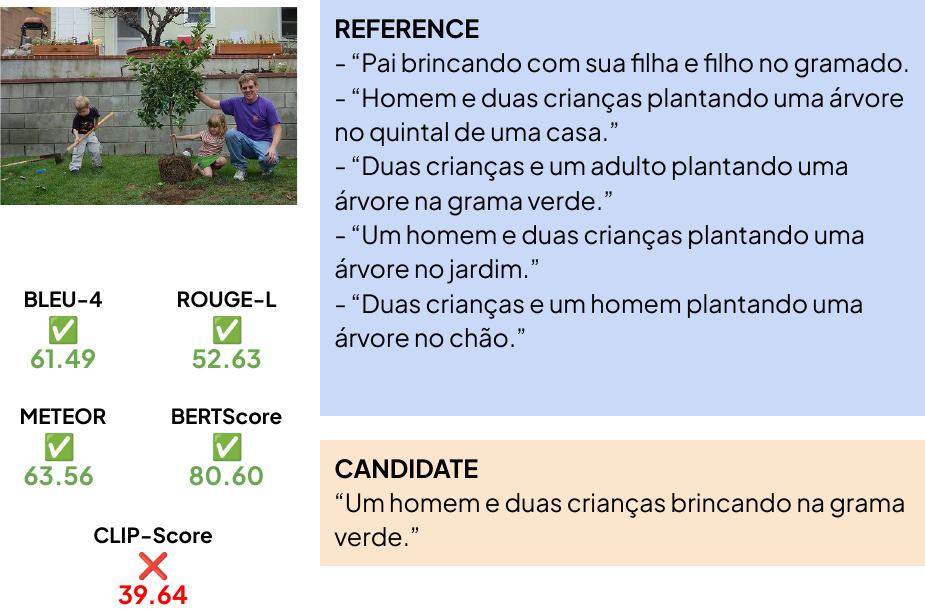}
    \caption{Case 4: Example from Flickr30K Native with a lower CLIP-Score compared to the uni-modal metrics.}
    \label{fig:worst-clip-native}
\end{figure}

Figures~\ref{fig:best-clip-translated} to~\ref{fig:worst-clip-native} illustrate four test cases under the same-source setup, presenting the reference captions, the generated captions (candidates) produced by Swin-DistilBERTimbau, and the corresponding performance metrics.

\paragraph{Case 1.} Figure~\ref{fig:best-clip-translated} shows a case from Flickr30K Translated with comparatively high performance on CLIP-Score given the uni-modal metrics. The generated caption was ``Três meninos grelhando carne.'' (``Three boys grilling meat.''). While the elements in the generated caption are correct, it lacks the level of detail present in the reference captions, and its length does not align with the reference captions either. It was observed a counting issue: the model inferred three people rather than four (real situation). Nonetheless, some could argue that there are ``at least'' three people in the image; therefore, the caption could be regarded as corrected. Finally, some words in the generated caption do not appear together in a single reference caption, but spread across distinct reference captions. For instance, the word ``meninos'' does not occur together with ``carne'' or words similar to ``grelhando'' (e.g., ``grelha''), affecting the performance of the reference-based (uni-modal) metrics.

\paragraph{Case 2.} Figure~\ref{fig:worst-clip-translated} depicts a case from Flickr30K Translated with poor performance on CLIP-Score compared to the uni-modal metrics, with the generated caption ``Três pessoas estão sentadas em um banco.'' (``Three people are sitting on a bench.''). 
In this case, the generated caption matches one of the reference captions in size and details, enabling high scores in reference-based metrics, while CLIP-Score scores are relatively lower. 
It is worth pointing out that the image is cropped around the middle of the people's bodies and the back support, which may impact the CLIP-Score evaluation. 
This cropping limits the availability of visual information, such as body positions and background elements, potentially affecting the alignment between image features and text.
Even disregarding these points, the reference captions remain more informative and descriptive than the candidate caption, containing descriptions about the matter of the bench (e.g., wood) and the ad with black letters on the back wall. 
The reference captions also carry some uncertain information, such as the location (e.g., shopping) and some elements in the image (e.g., escalator).

\paragraph{Case 3.} Figure~\ref{fig:best-clip-native} shows a case from Flickr30K Native with comparatively high performance on CLIP-Score given the uni-modal metrics. The generated caption is ``Um grupo de homens uniformizados tocando tambor.'' (``A group of uniformed men playing a drum.''). We identified the same issue in Case 1 (Figure~\ref{fig:best-clip-translated}), where the generated caption, despite its correctness, does not reach the same level of detail and length as the reference captions. There is some uncertainty on the expression ``Um grupo de homens'' (``a group of men'') because there may be a woman in the group. In this regard, other generic words (e.g., people) could provide a most secure alternative to gender-specific terms.

\paragraph{Case 4.} Figure~\ref{fig:worst-clip-native} shows a case from Flickr30K Native with poor performance on CLIP-Score in view of the uni-modal metrics. The generated caption was ``Um homem e duas crianças brincando na grama verde.'' (``A man and two children playing on the green grass.''), reaching a comparable level of detail and length to the reference captions. Only one reference caption includes the word ``brincando'' (``playing''), but the reference-based metrics achieved high values regardless of this consideration.

\subsubsection{Attention Maps} 

The attention in the cross-attention layer was utilized to associate text tokens with the image's attention matrix. For this analysis, the same cases in the previous section were revisited. To produce visualization at a word level, token-level attention was converted into word-level attention by summing the attention maps of the tokens. For instance, the word ``sentada'' (``sitting'') is the fusion of the tokens ``sent'' and ``\#\#ada''. The double hashtag in the token signifies a connection to the previous token, forming a complete word. The attention visualization represents a sequence of word-image pairs (or simply word-image), where whitened rectangles highlight areas with high attention values, while darkened areas indicate low attention values.

\begin{figure*}[ht]
    \centering
    \includegraphics[
        width=.9\linewidth,
        trim={5.5cm 1.2cm 4cm 0.9cm},
        clip
    ]{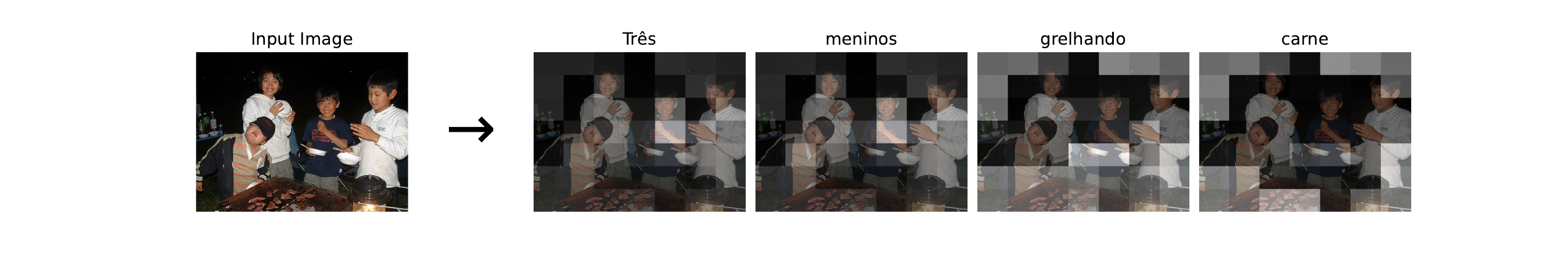}
    \caption{Attention visualization for Case 1 (Figure~\ref{fig:best-clip-translated}). Generated caption: ``Três meninos grelhando carne.'' (``Three boys grilling meat.'').}
    \label{fig:best-clip-maps-translated}
\end{figure*}

\paragraph{Case 1.} In Figure~\ref{fig:best-clip-maps-translated}, the word-images for ``grelhando'' (``grilling'') and ``carne'' (``meat'') coherently highlight areas with plates, grill, and some meat. However, other regions unrelated to these concepts are also highlighted, like the night sky. Some noise is observed in the inference of the number of people in the scene, which is evidenced by whitened pixels outside the expected regions and high focus on the child's arm.

\begin{figure*}[ht]
    \centering
    \includegraphics[
        width=.9\linewidth,
        trim={5.5cm 3.3cm 4cm 3.1cm},
        clip
    ]{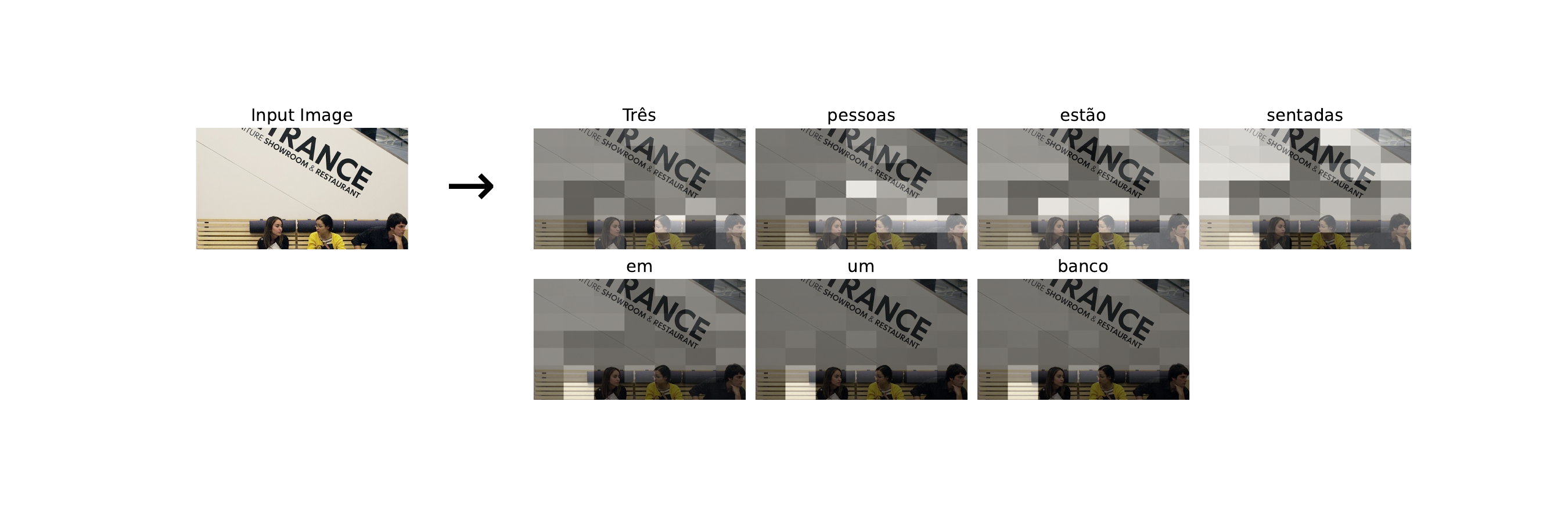}
    \caption{Attention visualization for Case 2 (Figure~\ref{fig:worst-clip-translated}). Generated caption: ``Três pessoas estão sentadas em um banco.'' (``Three people are sitting on a bench.'').}
    \label{fig:worst-clip-maps-translated}
\end{figure*}

\paragraph{Case 2.} In Figure~\ref{fig:worst-clip-maps-translated}, the model attention highlights areas with people, including when it outputs ``Três'', inferring the number of people in the scene. In this example, we can also observe noise-whitening regions that do not contain word-related elements, such as in the words ``pessoas'' (``people'') and ``sentadas'' (``sitting''). Despite the correct highlight for the word ``banco'' (``bench''), the previous pair ``em um'' (``on a'') -- stop words connected to ``banco'' -- yielded similar attention.

\begin{figure*}[ht]
    \centering
    \includegraphics[
        width=.9\linewidth,
        trim={5.5cm 2.2cm 4cm 2cm},
        clip
    ]{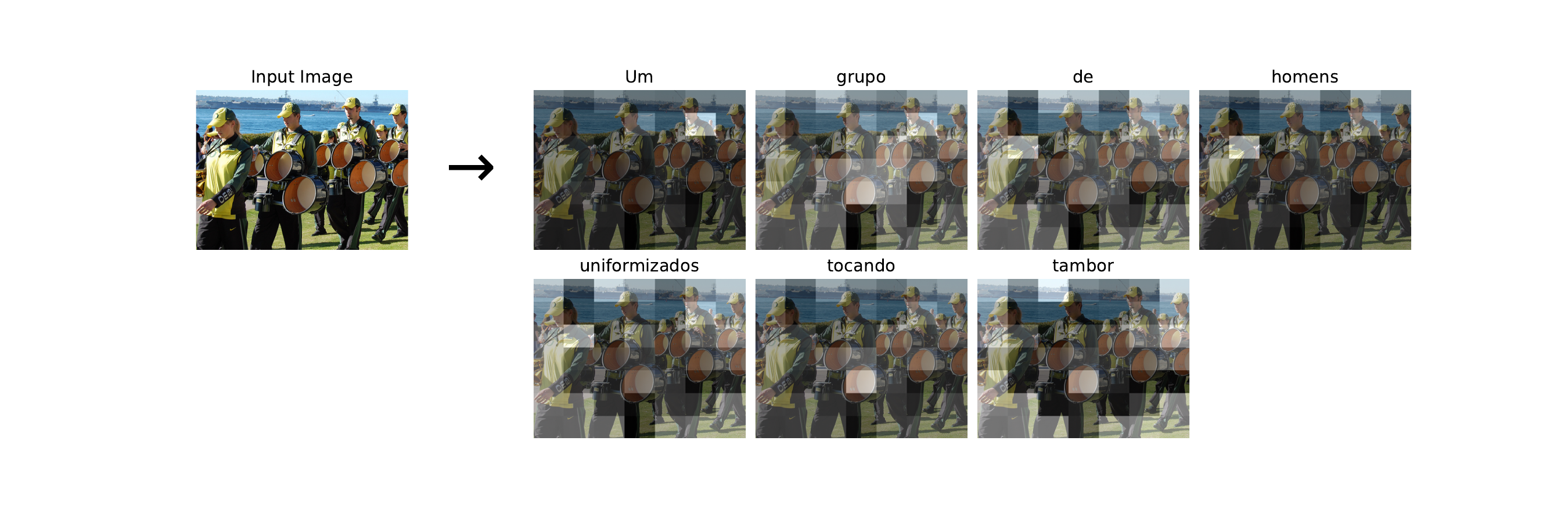}
    \caption{Attention visualization for Case 3 (Figure~\ref{fig:best-clip-native}). Generated caption: ``Um grupo de homens uniformizados tocando tambor.'' (``A group of uniformed men playing a drum.'').}
    \label{fig:best-clip-maps-native}
\end{figure*}

\paragraph{Case 3.} In Figure~\ref{fig:best-clip-maps-native}, the word-image pair ``homens'' (``men'') showed attention toward the people. In this case, the model exhibited a preference for a gender-specific term, a behavior that is a potential source of errors. Previous work, such as \citet{Girota2023GenderBiasIC}, also made such observations, demonstrating that captioning models perpetuate some society gender-related biases. The model correctly highlights the regions for the words ``uniformizados'' (``uniformed'') and ``tambor'' (``drum'').  

\begin{figure*}[ht]
    \centering
    \includegraphics[
        width=.9\linewidth,
        trim={5.5cm 4.1cm 4cm 4cm},
        clip
    ]{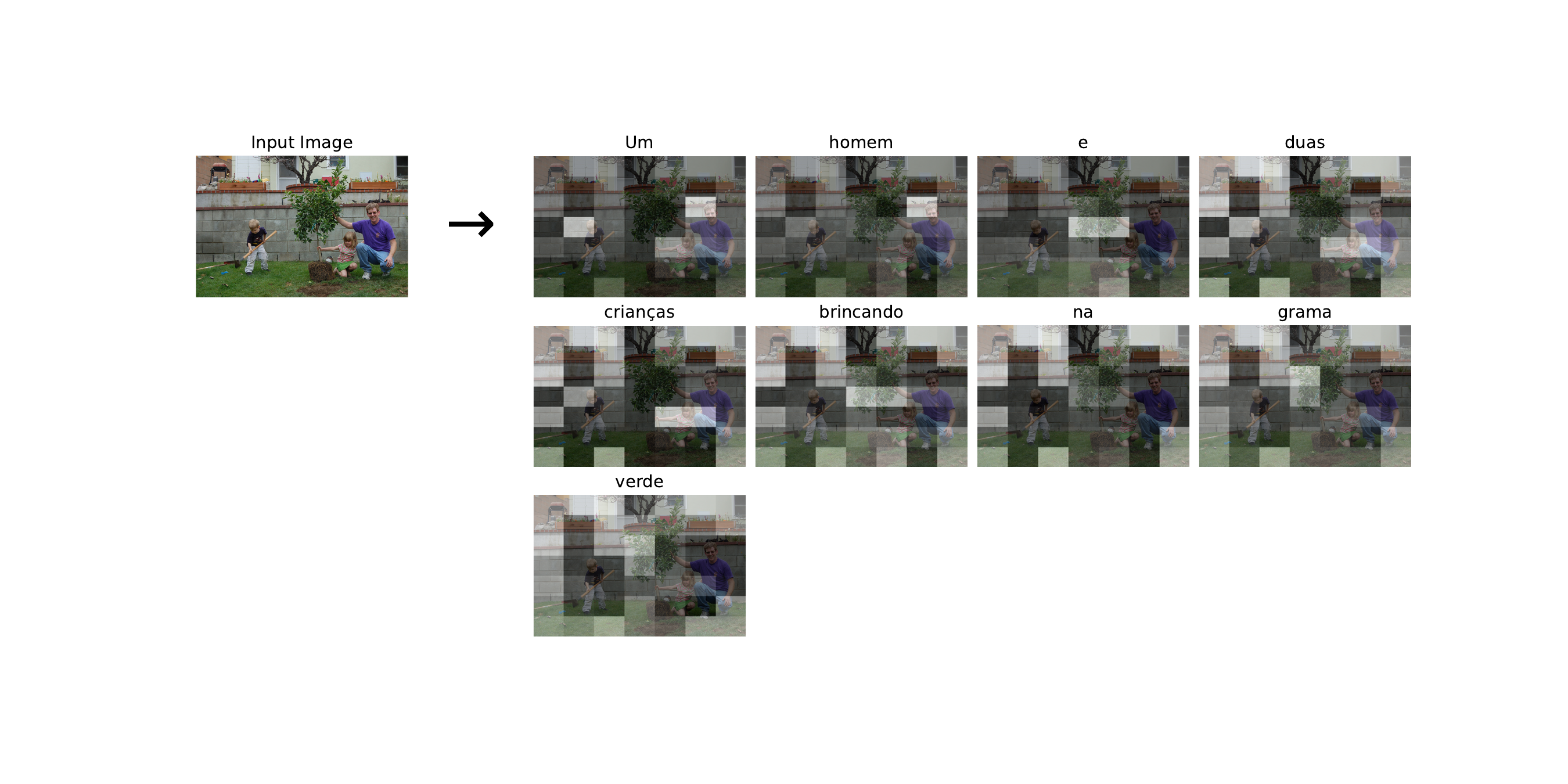}
    \caption{Attention visualization for Case 4 (Figure~\ref{fig:worst-clip-native}). Generated caption: ``Um homem e duas crianças brincando na grama verde.'' (``A man and two children playing on the green grass.'').}
    \label{fig:worst-clip-maps-native}
\end{figure*}

\paragraph{Case 4.} In Figure~\ref{fig:worst-clip-maps-native}, the most coherent maps are those related to the words ``Um homem'' (``A man''), ``duas crianças'' (``two children''), and ``grama verde'' (``green grass''). 
Initially, the model highlights the man on the map with a lighter white rectangle, while the children are marked with slightly darker white rectangles. In the sequence, the attention focuses on the children, but with little more attention to the little girl besides the man. The last two words, ``grama verde'' (``green grass''), have maps highlighting the surroundings of the people in the image.

\section{Concluding Remarks} \label{sec:conclusion}

This work evaluated the efficacy of Transformer-based VED models for Brazilian Portuguese image captioning under two main scenarios: (\textit{i})~training and testing on the same dataset (native or translated), and (\textit{ii})~cross-context between native and translated captions. 
Furthermore, we analyzed fine-tuned transformer-based VED models with sizes up to 240 million parameters compared to VLMs composed of more than 1.5 billion parameters.

Our results show that Swin encoders consistently outperform DeiT and ViT in text-centric and cross-modal metrics in Translated and Native scenarios, even when applying source-changing. 
Models based on GPorTuguese-2 exhibited the most pronounced performance degradation in the Flickr-Native test set when the training source shifted from native to translated.
Among the evaluation metrics, CLIP-Score was the most resilient to this change, as it operates independently of reference captions. In contrast, CIDEr exhibited the greatest performance decline.

Among the VLMs, ViTucano models were the only models that performed comparably with VED models in evaluation metrics. 
The remainder of the VLMs underperformed the VED models in reference-based metrics, indicating that the generated captions of VLMs have lower lexical matching with reference captions than the captions generated by the VED models. 
Nonetheless, the VLMs achieved higher CLIP-Score than VED models, except for PaliGemma and Phi-3V, indicating higher image-text embedding congruence than VED models.
Notably, native pre-trained models like ViTucano outperform larger multilingual models in reference-based evaluation metrics, including LLaMa 3.2 Vision, GPT-4o, and GPT-4o mini.
Additionally, LLaMa 3.2 Vision, an open-source model with 11 billion parameters, surpasses complex closed-source models like GPT-4o in several evaluation metrics.
PaliGemma and Phi-3 Vision presented the lowest performance.

The qualitative results analysis is supported by the addition of CLIP-Score and the attention maps visualization.
CLIP-Score is useful for evaluation in scenarios where traditional metrics fail to consider the correct information, such as cases where information is spread across the reference captions, making it difficult for traditional metrics to evaluate this information when it occurs together. 
In this situation, CLIP-Score does not replace traditional metrics for concisely evaluating captions, mainly when exact lexical matches occur.
Attention maps play an important role in model inference investigation, mainly to monitor whether the model is linking incoherent image regions with the target word. 
Some semantic connection mistakes were observed during the attention analysis, such as wrong attributions of gender, numerals, and adjectives due to erroneous region-word linking.

\paragraph{Limitations and Future Directions.}

This study focuses primarily on the Flickr30K dataset, which may limit generalizability to other image styles or cultural contexts. 
Additionally, while we compare native vs. machine-translated data, the quality of translation may vary by domain, length, or style, and further investigation into advanced translation or data augmentation strategies could prove beneficial.
The implicit bias of the annotators of these datasets is an issue since the two groups of annotators can diverge in point of view, attention to detail, language vices, and mother-language vocabulary. Some terms in English can not be easily translated to Portuguese while maintaining their original characteristics. 
Finally, we did not explore the impact of very large multi-modal transformers that have shown recent promise; adapting such models to Brazilian Portuguese might further bridge domain gaps.
Future work will expand these experiments to other state-of-the-art VLMs and investigate more sophisticated fine-tuning or adapter approaches, particularly for low-resource languages. 
Furthermore, we aim to expand the investigation of VLMs for Brazilian Portuguese IC by incorporating additional native datasets, such as \#PraCegoVer, and further translated sources.
The application of advanced statistical analysis over the evaluation metrics \citep{kilickaya2017re}, prompt engineering techniques to enhance VLMs performance \citep{wang2023controllable}, and additional evaluation metrics, such as LLM-as-a-judge \citep{chan2023clair}, are valuable additions to increase the present study.
By addressing these limitations, we aim to bring more inclusive and effective image captioning solutions for Brazilian Portuguese speakers and beyond.

\bibliographystyle{apalike}
\bibliography{references}

@Article{santos2022mdpi,
    AUTHOR = {dos Santos, Gabriel Oliveira and Colombini, Esther Luna and Avila, Sandra},
    TITLE = {{\#}PraCegoVer: A Large Dataset for Image Captioning in Portuguese},
    JOURNAL = {Data},
    VOLUME = {7},
    YEAR = {2022},
    NUMBER = {2},
    ARTICLE-NUMBER = {13},
    URL = {https://www.mdpi.com/2306-5729/7/2/13},
    ISSN = {2306-5729},
    DOI = {10.3390/data7020013}
}

@article{dosovitskiy2020image,
  title={An image is worth 16x16 words: Transformers for image recognition at scale},
  author={Dosovitskiy, Alexey and Beyer, Lucas and Kolesnikov, Alexander and Weissenborn, Dirk and Zhai, Xiaohua and Unterthiner, Thomas and Dehghani, Mostafa and Minderer, Matthias and Heigold, Georg and Gelly, Sylvain and others},
  journal={arXiv preprint arXiv:2010.11929},
  year={2020},
  doi={10.48550/arXiv.2010.11929}
}

@inproceedings{liu2021swin,
  title={Swin transformer: Hierarchical vision transformer using shifted windows},
  author={Liu, Ze and Lin, Yutong and Cao, Yue and Hu, Han and Wei, Yixuan and Zhang, Zheng and Lin, Stephen and Guo, Baining},
  booktitle={Proceedings of the IEEE/CVF international conference on computer vision},
  pages={10012--10022},
  year={2021},
  doi={10.1109/ICCV48922.2021.00986}
}

@inproceedings{papineni2002bleu,
  title={Bleu: a method for automatic evaluation of machine translation},
  author={Papineni, Kishore and Roukos, Salim and Ward, Todd and Zhu, Wei-Jing},
  booktitle={Proceedings of the 40th annual meeting of the Association for Computational Linguistics},
  pages={311--318},
  year={2002},
  doi={10.3115/1073083.1073135}
}

@inproceedings{banerjee2005meteor,
  title={METEOR: An automatic metric for MT evaluation with improved correlation with human judgments},
  author={Banerjee, Satanjeev and Lavie, Alon},
  booktitle={Proceedings of the acl workshop on intrinsic and extrinsic evaluation measures for machine translation and/or summarization},
  pages={65--72},
  year={2005}
}

@inproceedings{lin2004rouge,
  title={Rouge: A package for automatic evaluation of summaries},
  author={Lin, Chin-Yew},
  booktitle={Text summarization branches out},
  pages={74--81},
  year={2004}
}

@article{zhang2019bertscore,
  title={Bertscore: Evaluating text generation with bert},
  author={Zhang, Tianyi and Kishore, Varsha and Wu, Felix and Weinberger, Kilian Q and Artzi, Yoav},
  journal={arXiv preprint arXiv:1904.09675},
  year={2019},
  doi={10.48550/arXiv.1904.09675}
}

@inproceedings{vedantam2015cider,
  title={Cider: Consensus-based image description evaluation},
  author={Vedantam, Ramakrishna and Lawrence Zitnick, C and Parikh, Devi},
  booktitle={Proceedings of the IEEE conference on computer vision and pattern recognition},
  pages={4566--4575},
  year={2015},
  doi={10.1109/CVPR.2015.7299087}
}

@article{stefanini2022show,
  title={From show to tell: A survey on deep learning-based image captioning},
  author={Stefanini, Matteo and Cornia, Marcella and Baraldi, Lorenzo and Cascianelli, Silvia and Fiameni, Giuseppe and Cucchiara, Rita},
  journal={IEEE transactions on pattern analysis and machine intelligence},
  volume={45},
  number={1},
  pages={539--559},
  year={2022},
  publisher={IEEE},
  doi={10.1109/TPAMI.2022.3148210}
}

@article{sharma2023comprehensive,
  title={A comprehensive survey on image captioning: from handcrafted to deep learning-based techniques, a taxonomy and open research issues},
  author={Sharma, Himanshu and Padha, Devanand},
  journal={Artificial Intelligence Review},
  pages={1--43},
  year={2023},
  publisher={Springer},
  doi={10.1007/s10462-023-10488-2}
}

@inproceedings{vinyals2015show,
  title={Show and tell: A neural image caption generator},
  author={Vinyals, Oriol and Toshev, Alexander and Bengio, Samy and Erhan, Dumitru},
  booktitle={Proceedings of the IEEE conference on computer vision and pattern recognition},
  pages={3156--3164},
  year={2015},
  doi={10.1109/CVPR.2015.7298935}
}

@inproceedings{souza2020bertimbau,
  title={BERTimbau: pretrained BERT models for Brazilian Portuguese},
  author={Souza, F{\'a}bio and Nogueira, Rodrigo and Lotufo, Roberto},
  booktitle={Intelligent Systems: 9th Brazilian Conference, BRACIS 2020, Rio Grande, Brazil, October 20--23, 2020, Proceedings, Part I 9},
  pages={403--417},
  year={2020},
  organization={Springer},
  doi={10.1007/978-3-030-61377-8\_28}
}

@article{de2024image,
  title={Image captioning for Brazilian Portuguese using GRIT model},
  author={de Alencar, Rafael Silva and Casta{\~n}eda, William Alberto Cruz and Amadeus, Marcellus},
  journal={arXiv preprint arXiv:2402.05106},
  year={2024},
  doi={10.48550/arXiv.2402.05106}
}

@inproceedings{gondim2022towards,
  title={Towards Image Captioning for the Portuguese Language: Evaluation on a Translated Dataset.},
  author={Gondim, Jo{\~a}o and Claro, Daniela Barreiro and Souza, Marlo},
  booktitle={ICEIS (1)},
  pages={384--393},
  year={2022},
  doi={10.5220/001108000000317}
}

@article{silva2022towards,
  title={Towards transfer learning techniques—BERT, DistilBERT, BERTimbau, and DistilBERTimbau for automatic text classification from different languages: a case study},
  author={Silva Barbon, Rafael and Akabane, Ademar Takeo},
  journal={Sensors},
  volume={22},
  number={21},
  pages={8184},
  year={2022},
  publisher={MDPI},
 doi={10.3390/s22218184}
}

@misc {adalberto_ferreira_barbosa_junior_2024,
    author       = { {Adalberto Ferreira Barbosa Junior} },
    title        = { distilbert-portuguese-cased (Revision df1fa7a) },
    year         = 2024,
    url          = { https://huggingface.co/adalbertojunior/distilbert-portuguese-cased },
    doi          = { 10.57967/hf/3041 },
    publisher    = { Hugging Face }
}

@inproceedings{touvron2021training,
  title={Training data-efficient image transformers \& distillation through attention},
  author={Touvron, Hugo and Cord, Matthieu and Douze, Matthijs and Massa, Francisco and Sablayrolles, Alexandre and J{\'e}gou, Herv{\'e}},
  booktitle={International conference on machine learning},
  pages={10347--10357},
  year={2021},
  organization={PMLR}
}

@inproceedings{pierre2020gpt2smallportuguese,
  title={GPorTuguese-2 (Portuguese GPT-2 small): a Language Model for Portuguese text generation (and more NLP tasks...)},
  author={Pierre Guillou},
  year={2020},
  url={https://huggingface.co/pierreguillou/gpt2-small-portuguese}
}

@inproceedings{viridiano2024framed,
  title={Framed Multi30K: A Frame-Based Multimodal-Multilingual Dataset},
  author={Viridiano, Marcelo and Lorenzi, Arthur and Torrent, Tiago Timponi and Matos, Ely E and Pagano, Adriana S and Sigiliano, Nat{\'a}lia Sathler and Gamonal, Maucha and de Andrade Abreu, Helen and Dutra, L{\'\i}via Vicente and Samagaio, Mairon and others},
  booktitle={Proceedings of the 2024 Joint International Conference on Computational Linguistics, Language Resources and Evaluation (LREC-COLING 2024)},
  pages={7438--7449},
  year={2024}
}

@inproceedings{dos2023capivara,
  title={CAPIVARA: Cost-Efficient Approach for Improving Multilingual CLIP Performance on Low-Resource Languages},
  author={dos Santos, Gabriel Oliveira and Moreira, Diego Alysson Braga and Ferreira, Alef Iury and Silva, Jhessica and Pereira, Luiz and Bueno, Pedro and Sousa, Thiago and Maia, Helena and Da Silva, N{\'a}dia and Colombini, Esther and others},
  booktitle={Proceedings of the 3rd Workshop on Multi-lingual Representation Learning (MRL)},
  pages={184--207},
  year={2023},
  doi={10.18653/v1/2023.mrl-1.15}
}

@inproceedings{hessel2021clipscore,
  title={CLIPScore: A Reference-free Evaluation Metric for Image Captioning},
  author={Hessel, Jack and Holtzman, Ari and Forbes, Maxwell and Le Bras, Ronan and Choi, Yejin},
  booktitle={Proceedings of the 2021 Conference on Empirical Methods in Natural Language Processing},
  pages={7514--7528},
  year={2021},
  doi={10.18653/v1/2021.emnlp-main.595}
}

@article{ghandi2023deep,
  title={Deep learning approaches on image captioning: A review},
  author={Ghandi, Taraneh and Pourreza, Hamidreza and Mahyar, Hamidreza},
  journal={ACM Computing Surveys},
  volume={56},
  number={3},
  pages={1--39},
  year={2023},
  publisher={ACM New York, NY},
  doi={10.1145/3617592}
}

@inproceedings{jnaini2024synergy,
  title={Synergy of GPT-3 Summarization and Vision-Encoder-Decoder for Chest X-ray Captioning},
  author={Jnaini, Abdellah and Shirazi, Hossein and Homayouni, Hajar},
  booktitle={2024 IEEE Canadian Conference on Electrical and Computer Engineering (CCECE)},
  pages={476--482},
  year={2024},
  organization={IEEE},
  doi={10.1109/CCECE59415.2024.10667261}
}

@inproceedings{abdelaal2024image,
  title={Image Captioning Using Vision Encoder Decoder Model},
  author={Abdelaal, Ahmad and ELshafey, Nadeen Farid and Abdalah, Nadine Walid and Shaaban, Nouran Hady and Okasha, Sama Ahmed and Yasser, Tawfik and Fathi, Mostafa and Fouad, Khaled M and Abdelbaky, Ibrahim},
  booktitle={2024 International Conference on Machine Intelligence and Smart Innovation (ICMISI)},
  pages={101--106},
  year={2024},
  organization={IEEE},
  doi={10.1109/ICMISI61517.2024.10580628}
}

@INPROCEEDINGS{ishan2023bengali,
  author={Ishan, Tajrian Islam and Al Noman, Abdullah and Rokib, Raisa and Masum, Mustavi Ibne and Ahmed, Sifat and Shah, Faisal Muhammad},
  booktitle={2023 26th International Conference on Computer and Information Technology (ICCIT)}, 
  title={Bengali Image Captioning Using Vision Encoder-Decoder Model}, 
  year={2023},
  volume={},
  number={},
  pages={1-6},
  doi={10.1109/ICCIT60459.2023.10441125}}

@inproceedings{bromonschenkel2024comparative,
  title={A Comparative Evaluation of Transformer-Based Vision Encoder-Decoder Models for Brazilian Portuguese Image Captioning},
  author={Bromonschenkel, Gabriel and Oliveira, Hil{\'a}rio and Paix{\~a}o, Thiago M},
  booktitle={2024 37th SIBGRAPI Conference on Graphics, Patterns and Images (SIBGRAPI)},
  pages={1--6},
  year={2024},
  organization={IEEE},
  doi={10.1109/SIBGRAPI62404.2024.10716325}
}

@article{liu2021cptr,
  title={Cptr: Full transformer network for image captioning},
  author={Liu, Wei and Chen, Sihan and Guo, Longteng and Zhu, Xinxin and Liu, Jing},
  journal={arXiv preprint arXiv:2101.10804},
  year={2021},
  doi={10.48550/arXiv.2101.10804}
}

@misc{Correa20204VITucano,
    author={Corr{\^e}a, Nicholas Kluge and Sen, Aniket and Falk, Sophia and Fatimah, Shiza},
    title={{ViTucano: A Portuguese Vision Assitant}},
    year=2024,
    howpublished = {\url{https://huggingface.co/TucanoBR}},
}

@article{Beyer2024Paligemma,
  title={Paligemma: A versatile 3b vlm for transfer},
  author={Beyer, Lucas and Steiner, Andreas and Pinto, Andr{\'e} Susano and Kolesnikov, Alexander and Wang, Xiao and Salz, Daniel and Neumann, Maxim and Alabdulmohsin, Ibrahim and Tschannen, Michael and Bugliarello, Emanuele and others},
  journal={arXiv preprint arXiv:2407.07726},
  year={2024},
  doi={10.48550/arXiv.2407.07726}
}

@article{Abdin2024Phi3,
  title={Phi-3 technical report: A highly capable language model locally on your phone},
  author={Abdin, Marah and Aneja, Jyoti and Awadalla, Hany and Awadallah, Ahmed and Awan, Ammar Ahmad and Bach, Nguyen and Bahree, Amit and Bakhtiari, Arash and Bao, Jianmin and Behl, Harkirat and others},
  journal={arXiv preprint arXiv:2404.14219},
  year={2024},
  doi={10.48550/arXiv.2404.14219}
}

@article{Dubey2024Llama,
  title={The llama 3 herd of models},
  author={Dubey, Abhimanyu and Jauhri, Abhinav and Pandey, Abhinav and Kadian, Abhishek and Al-Dahle, Ahmad and Letman, Aiesha and Mathur, Akhil and Schelten, Alan and Yang, Amy and Fan, Angela and others},
  journal={arXiv preprint arXiv:2407.21783},
  year={2024},
  doi={10.48550/arXiv.2407.21783}
}

@inproceedings{Girota2023GenderBiasIC,
  title={Model-agnostic gender debiased image captioning},
  author={Hirota, Yusuke and Nakashima, Yuta and Garcia, Noa},
  booktitle={Proceedings of the IEEE/CVF Conference on Computer Vision and Pattern Recognition},
  pages={15191--15200},
  year={2023},
  doi={10.1109/CVPR52729.2023.01458}
}

@inproceedings{kilickaya2017re,
  title={Re-evaluating Automatic Metrics for Image Captioning},
  author={Kilickaya, Mert and Erdem, Aykut and Ikizler-Cinbis, Nazli and Erdem, Erkut},
  booktitle={Proceedings of the 15th Conference of the European Chapter of the Association for Computational Linguistics: Volume 1, Long Papers},
  pages={199--209},
  year={2017}
}

@inproceedings{chan2023clair,
  title={CLAIR: Evaluating Image Captions with Large Language Models},
  author={Chan, David and Petryk, Suzanne and Gonzalez, Joseph and Darrell, Trevor and Canny, John},
  booktitle={Proceedings of the 2023 Conference on Empirical Methods in Natural Language Processing},
  pages={13638--13646},
  year={2023},
  doi={10.18653/v1/2023.emnlp-main.841}
}

@inproceedings{wang2023controllable,
  title={Controllable image captioning via prompting},
  author={Wang, Ning and Xie, Jiahao and Wu, Jihao and Jia, Mingbo and Li, Linlin},
  booktitle={Proceedings of the AAAI Conference on Artificial Intelligence},
  volume={37},
  number={2},
  pages={2617--2625},
  year={2023},
  doi={10.1609/aaai.v37i2.25360}
}

@article{hurst2024gpt,
  title={Gpt-4o system card},
  author={Hurst, Aaron and Lerer, Adam and Goucher, Adam P and Perelman, Adam and Ramesh, Aditya and Clark, Aidan and Ostrow, AJ and Welihinda, Akila and Hayes, Alan and Radford, Alec and others},
  journal={arXiv preprint arXiv:2410.21276},
  year={2024},
  doi={10.48550/arXiv.2410.21276}
}

\end{document}